\documentclass[11pt]{article}

\usepackage[final]{acl}

\usepackage{times}
\usepackage{latexsym}

\usepackage[T1]{fontenc}
\usepackage[utf8]{inputenc}

\usepackage{microtype}

\usepackage{inconsolata}

\usepackage{graphicx}
\usepackage{color}
\usepackage{booktabs}
\usepackage{tabularx}
\usepackage{adjustbox}
\usepackage{pdflscape}
\usepackage{forest}

\usepackage[most]{tcolorbox}
\usepackage{cleveref}

\usepackage{xcolor}
\usepackage[normalem]{ulem}

\newif\ifdraft
\drafttrue          % <<< set \draftfalse for submission
\draftfalse

\usepackage[normalem]{ulem}
\usepackage{xcolor}

\newcommand{\yr}[2]{\ifdraft\sout{#1}\,\textcolor{violet}{[#2]}\else#2\fi}

\title{The Roots of Performance Disparity in Multilingual Language Models:\\ Intrinsic Modeling Difficulty or Design Choices?}

\author{
 \textbf{Chen Shani\textsuperscript{1}},
 \textbf{Yuval Reif\textsuperscript{2}},
 \textbf{Nathan Roll\textsuperscript{1}},
 \textbf{Dan Jurafsky\textsuperscript{1}},
 \textbf{Ekaterina Shutova\textsuperscript{3}}
\\
 \textsuperscript{1}Stanford University,
 \textsuperscript{2}The Hebrew University of Jerusalem,
 \textsuperscript{3}University of Amsterdam
}

\begin{document}
\maketitle
\begin{abstract}

Multilingual language models (LMs) promise broader NLP access, yet current systems deliver uneven performance across the world’s languages. This survey examines why these gaps persist and whether they reflect intrinsic linguistic difficulty or modeling artifacts. We organize the literature around two questions: do linguistic disparities arise from representation and allocation choices (e.g., tokenization, encoding, data exposure, parameter sharing) rather than inherent complexity; and which design choices mitigate inequities across typologically diverse languages. We review linguistic features, such as orthography, morphology, lexical diversity, syntax, information density, and typological distance, linking each to concrete modeling mechanisms. Gaps often shrink when segmentation, encoding, and data exposure are normalized, suggesting much apparent difficulty stems from current modeling choices. We synthesize these insights into design recommendations for tokenization, sampling, architectures, and evaluation to support more balanced multilingual LMs. 
\end{abstract}

\section{Introduction}

Multilingual LMs have expanded NLP's reach by enabling a single model to perform tasks across many languages. They are pretrained on text from hundreds of languages, sharing parameters and representations~\citep{devlin-etal-2019-bert, conneau-etal-2020-unsupervised, scao2022bloom, imanigooghari-etal-2023-glot500, aya-expanse}. This  enables cross-lingual transfer, where patterns learned in one language improve performance in others~\citep{pires2019multilingual, wu2019emerging, lauscher2020zero, malkin-etal-2022-balanced, blevins2024breaking}. Despite these advantages, persistent performance disparities across languages limit the practical reach of multilingual models~\cite{wang2025uncovering, ghosh2025survey}.

These disparities systematically follow cross-linguistic patterns: higher-resource languages and those structurally similar to dominant training languages generally perform better than low-resource or typologically distant ones~\cite{zhao2025comprehensive, akindotuni2025resource}. The disparities often persist even with large-scale pretraining, suggesting that scaling alone cannot ensure equitable performance~\citep{hoffmann-etal-2022-chinchilla, he-etal-2025-multilingual-scaling-laws}. This raises a central question: \textcolor{red!70!black}{\textbf{are some languages inherently harder to model, or do performance gaps reflect engineering artifacts and design choices?}}

We review how linguistic structure interacts with multilingual design choices to shape performance gaps via two questions: whether disparities stem from intrinsic difficulty or modeling artifacts (e.g., tokenization, data allocation, shared-parameter interference); and which design choices mitigate inequities. We consolidate our findings into a set of recommendations for tokenization, data sampling, model architectures, and evaluation, highlighting where evaluations confound learnability with tokenization or encoding artifacts (\Cref{tab:mechanism_map}).

Our synthesis suggests that cross-linguistic gaps rarely reflect intrinsic modeling complexity. Instead, they arise via three mechanisms: (1) shared-parameter training induces negative transfer when typological diversity exceeds effective capacity~\citep{pfeiffer2022lifting, blevins2024breaking, chang2024multilinguality}; (2) tokenization and encoding fragment words or penalize byte-heavy scripts, inflating sequence length without added meaning~\citep{rust2021good, arnett2024bit, lundin2025tokentax, land2025script}; and (3) data sampling and evaluation misrepresent semantic exposure. Gaps shrink when normalizing segmentation, encoding, and exposure or explicitly allocating capacity, indicating that difficulty stems from modeling choices.

This first systematic review of cross-linguistic modeling difficulty research offers practical design recommendations for multilingual LMs to achieve balanced performance across diverse languages.

% \cnote{add the checklist?}

\begin{table*}[t]
\centering
\footnotesize
\begin{adjustbox}{max width=\textwidth}
\begin{tabular}{p{4.0cm} p{4.2cm} p{4.2cm} p{4.4cm}}
\toprule
\textbf{Linguistic Factor} &
\textbf{Observed Artifact} &
\yr{\textbf{Mechanism (Technical Cause)}}{\textbf{Modeling Mechanism}} &
\textbf{Design Levers} \\
\midrule

Orthography and encoding granularity (\S\ref{sec:orthography}) &
Encoding inefficiency (byte premium); reduced effective exposure under fixed budgets;  inconsistent written signal &
UTF-8 byte-length asymmetries inflate sequence length and reduce effective training signal for many non-Latin scripts &
Byte-normalized sampling; script-aware or language-adaptive tokenization; alternative encodings; tokenizer-free models (\S\ref{sec:tokenization},   \S\ref{sec:sampling}) \\
\addlinespace[2pt]

Morphology: productivity and compounding (\S\ref{sec:morphology}, \S\ref{sec:lexical}) &
Tokenization (over-segmentation, longer sequences); diluted training signal across surface forms &
BPE subwords misalign with morpheme boundaries, yielding inconsistent tokenization &
Morphology-aware tokenization; language-aware vocabularies (\S\ref{sec:tokenization}) \\
\addlinespace[2pt]

Information density and redundancy (\S\ref{sec:information}) &
Unequal semantic coverage under fixed token budgets &
Token-based budgeting allocates unequal information per unit of training, confounding cross-language comparisons &
Information-, byte- or morpheme-normalized sampling; adaptive scaling of effective exposure to language data (\S\ref{sec:sampling}, \S\ref{sec:balanced_pretraining}) \\
\addlinespace[2pt]

\yr{Typological distance and syntactic divergence (\S\ref{sec:typology}, \S\ref{sec:syntax})}{Typological and syntactic divergence (\S\ref{sec:typology}, \S\ref{sec:syntax})} &
Negative transfer under shared parameters; degraded syntactic generalization &
\yr{Gradient conflict and representation collapse in shared capacity}{Shared capacity induces gradient conflict and representation collapse when languages differ strongly in structure} &
Modular capacity and language adapters; typology-aware routing; controlled sharing (\S\ref{sec:balanced_pretraining})%, \S\ref{sec:design_constraint}) 
\\
\addlinespace[2pt]

\yr{Tokenization and script variability in evaluation (\S\ref{sec:orthography}, \S\ref{sec:information})}{Evaluation sensitivity to tokenization and encoding (\S\ref{sec:orthography}, \S\ref{sec:information})} &
Perplexity comparability confounded by segmentation and byte length &
Subword-level metrics conflate segmentation decisions with predictability &
\yr{Character/byte/morpheme-level evaluation; typology-aware probes (\S\ref{sec:evaluation})}{Report character/morpheme-level metrics;  tokenization diagnostics and typology-aware probes (\S\ref{sec:evaluation})} \\

\bottomrule
\end{tabular}
\end{adjustbox}
\caption{Linking linguistic properties to multilingual modeling artifacts, mechanisms, and design levers; section references point to the supporting evidence discussed in the survey.}
\label{tab:mechanism_map}
\end{table*}

\section{Linguistic Properties}% Associated with Modeling Difficulty}

Human languages evolve under multiple, sometimes competing objectives, producing systematic trade-offs across morphology, syntax, and phonology~\citep{gibson2019efficiency}. Information may be densely packed within words or distributed across syntax; flexible word order can be balanced by overt marking such as case or agreement. These features preserve overall communicative efficiency despite wide typological diversity~\citep{gibson2019efficiency, lian-etal-2023-communication}. Human acquisition aligns with this view: children reliably acquire their ambient language, though the timing and difficulty of specific constructions vary by typology rather than defining a universal hierarchy of ``hard'' languages~\citep{slobin-1985-crosslinguistic-acquisition, berman-2014-crosslinguistic-child-language}.

We review linguistic properties associated with cross-linguistic performance variation, where \textit{learnability} refers to sample efficiency and predictive performance (perplexity, downstream accuracy). Drawing on NLP, computational linguistics, typology, and information theory, we show how these properties influence tokenization, data allocation, and architecture in multilingual LMs. Each subsection defines a property, summarizes evidence, and discusses factors affecting modeling success.

\subsection{Orthography}\label{sec:orthography}

\begin{tcolorbox}[colback=red!2!white,colframe=red!70!black]
% Orthography drives cross-linguistic disparities by skewing encoding efficiency across writing systems under equal byte budgets.
Orthography drives cross-linguistic disparities by shaping encoding efficiency and surface inconsistency across writing systems.
\end{tcolorbox}

Orthography concerns how linguistic content is represented in writing.
Writing systems differ in \emph{granularity}---whether symbols correspond roughly to phonemes, syllables, or morphemes---and in \emph{transparency}, or how predictably written forms map to sounds~\citep{wydell1999case, ziegler_goswami_2005_grain_size}.
In humans, these differences can influence the difficulty of literacy acquisition, rather than spoken-language learning itself~\citep{verhoeven2022universals,chang2020relationships}. Although children learn to read at different rates across languages~\citep{lai2024bigger,seymour2003foundation}, skilled adult reading is broadly similar across orthographies~\citep{schroeder2022eye,liversedge2016universality}.

Language models, however, acquire language directly from text, without prior phonological, lexical, or semantic knowledge. Thus, orthographic differences matter mostly as differences in representation efficiency---how meaning is encoded into bytes and tokens.
We focus on three consequences of orthography for modeling: encoding efficiency, vocabulary allocation in multilingual tokenizers, and the surface consistency of written forms.

First, writing systems differ in how much information they express per written symbol~\citep{orthography-information-content-2024} and in how those symbols are encoded under UTF-8~\citep{yergeau_2003_rfc3629_utf8}.
Alphabetic systems such as English distribute meaning across sequences of letters that roughly track phonological units, whereas logographic systems such as Chinese often express comparable content with fewer, denser characters~\citep{tan2001neural}. In abugidas such as Devanagari, characters are organized around consonantal bases, with vowels often expressed through attached diacritics~\citep{velayuthan-sarveswaran-2025-egalitarian}.
These differences lead to disparities at the byte level: English characters occupy one byte, Arabic characters two, and Chinese characters three; in Devanagari, what readers perceive as a single written unit may consist of a base consonant plus multiple diacritics, each separately encoded in three bytes~\citep{lemire2022transcoding,lavanya2005simple}.

This creates a \emph{byte premium}: under a fixed token or context budget, equal numbers of bytes do not correspond to equal amounts of linguistic content across languages~\citep{arnett2024bit}. For languages written in multi-byte scripts, the same content occupies longer encoded sequences, reducing effective exposure during training and shrinking the amount of text that fits within a context window~\citep{moon2025bitlevel}. The problem is not only that some scripts yield longer sequences, but that equal budgets systematically allocate unequal amounts of usable input across languages.

Second, orthography affects how efficiently tokenizers can build reusable units. Most modern language models operate on subword vocabularies learned from corpus statistics, often via byte-pair encoding~\citep[BPE;][]{sennrich-etal-2016-neural} starting from a byte-level vocabulary. Tokenization efficiency therefore depends not only on how much information each written symbol carries, but also on how easily recurring sequences can be merged into reusable units, which tends to disadvantage scripts whose characters are encoded with multiple bytes~\citep{sennrich-etal-2016-neural, zouhar-etal-2023-formal, kargaran-etal-2024-glotscript-resource}. Thus, even under the same vocabulary budget, this can produce large disparities in compression across languages.

In multilingual settings, shared-vocabulary tokenization can allocate capacity unevenly: high-resource Latin-script languages tend to receive larger and more informative subwords, while many non-Latin scripts are segmented into shorter fragments with less information per token~\citep{petrov2023language, ahia2023all}. Conversely, sharing or aligning scripts can improve transfer. For unseen or low-resource languages, transliteration into a script already well represented in the model can improve downstream performance~\citep{muller-etal-2021-unseen,moosa-etal-2023-transliteration}; it can also improve cross-lingual alignment, while recent work identifies script mismatch itself as a major barrier to cross-script knowledge transfer~\citep{moosa-etal-2023-transliteration,bandarkar2026large}.

Byte-level tokenization also introduces distortions specific to multi-byte scripts. Because BPE merges frequent byte sequences rather than linguistically meaningful units, tokens may split characters across byte boundaries or contain only partial UTF-8 sequences~\citep{firestone2025utf,jang-etal-2025-improbable,land2025script}. Such fragments need not correspond to phonological, morphological, or semantic structure. Unrelated symbols may therefore partially overlap in tokenization simply because they share bytes, while meaningful sub-character structure may be obscured when token boundaries fail to align with it~\citep{haslett-2025-tokenization}. A straightforward character-level vocabulary would avoid some of these artifacts, but in multilingual settings even a base vocabulary of Unicode characters would already be extremely large---on the order of 130k types~\citep{petrov2023language, ahia2023all}. Consequently, high information density per character does not necessarily yield equally efficient tokenization.

Third, orthography can shape the surface consistency of the training signal, because the same underlying content may appear in multiple written forms. This can arise from optional diacritics, as in Arabic and Hebrew~\citep{inoue-etal-2026-diacritics,gorman-pinter-2025-dont}; from widespread spelling inconsistencies~\citep{obeid-etal-2020-camel,adouane-etal-2019-normalising}; and from routine script alternation, such as Simplified versus Traditional Chinese~\citep{characterizing-bias2025-lyu} or South Asian languages commonly written in both native and Latin scripts~\citep{roark-etal-2020-processing}. As a result, models may observe the same meaning dispersed across several orthographic variants rather than concentrated in a single stable form.

Tokenizer-free models can reduce some of these disparities~\citep{pagnoni-etal-2025-byte,clark2022canine}, and recent methods such as MYTE mitigate script-specific penalties by introducing alternatives to UTF-8~\citep{limisiewicz2024myte, land2025script}. However, these approaches do not eliminate orthographic asymmetries: byte premiums still lengthen sequences even for tokenizer-free models, character granularity remains incomparable across scripts, and surface variation can still fragment training signal. Overall, orthography contributes to cross-linguistic disparities by making encoding efficiency unequal across writing systems and written signal more or less consistent before modeling even begins.

\subsection{Morphological Complexity}\label{sec:morphology}

\begin{tcolorbox}[colback=red!2!white,colframe=red!70!black]
Apparent performance gaps from rich morphology often stem from segmentation quality, vocabulary budget, and data allocation.
\end{tcolorbox}

Morphological complexity concerns how languages change or combine words to express distinctions such as tense, number, case, or word meaning, through processes such as inflection, derivation, and compounding~\citep{haspelmath_sims_2010_understanding_morphology}. Children ambiently learn the word-building patterns of their native language, although the pace of acquisition varies with the regularity and typology of the system~\citep{clark2017morphology}. For adult second-language learners, these patterns can remain difficult, especially when they differ substantially from those of the learner’s first language~\citep{ellis2022second}.
Perhaps because of this, morphology has often been assumed to increase language modeling difficulty~\citep{cotterell-etal-2018-equally, gerz-etal-2018-relation, park-etal-2021-morphology, mielke-etal-2019-kind}.

A common explanation is sparsity: morphologically rich languages realize each lexeme in many surface forms, lowering the frequency of individual forms and increasing the burden on models to generalize across paradigms even when the underlying rules are regular~\citep{park-etal-2021-morphology}. Early multilingual studies seemed to support this view: \citet{cotterell-etal-2018-equally} found that lower performance  correlates with morphological richness across 21 languages, and that this effect was largely removed by lemmatization. Similarly, \citet{gerz-etal-2018-relation} reported substantial morphology-related differences across 50 typologically diverse languages.

Later work, however, suggests that much of this apparent effect is not intrinsic to morphology itself. Instead, it is amplified by how current pipelines segment, encode, and sample morphologically complex languages~\citep{mielke-etal-2019-kind}. In particular, morphology-aware segmentation substantially reduces surprisal or performance gaps induced by standard BPE~\citep{park-etal-2021-morphology,mager2022bpe}, and analyses of WordPiece and BPE show that they often fail to preserve morpheme structure in languages with complex inflection or derivation~\citep{klein2020getting, lerner2025unlike}. 
Recent work further shows that once tokenization or effective exposure are controlled, morphological complexity is a much weaker predictor of LM performance than previously assumed~\citep{arnett2025language,asgari2025morphbpe,rust2021good}.

Taken together, these results suggest that morphology affects language modeling through three interacting mechanisms. First, tokenization determines whether recurring morphemes are preserved as reusable units or broken into arbitrary fragments; when morpheme boundaries are obscured, models can generalize less effectively across related word forms~\citep{mager2022bpe,park-etal-2021-morphology,bostrom2020byte,gazit-etal-2025-splintering}. Second, rich morphology can increase sequence cost when grammatical information is distributed across more fragmented token sequences, so the same content consumes more of the model's context, and equal token budgets result in less effective data exposure~\citep{arnett2024bit, foroutan2025paritybpe, asgari2025morphbpe}. Third, rich morphology can spread training signal across many low-frequency forms, while multilingual vocabulary learning may allocate less useful capacity to the morphemes needed to represent them, leaving less training signal for each individual form~\citep{reif2025vocab,park-etal-2021-morphology,rust2021good}.

Morphology is therefore best treated as an interaction effect rather than a uniform source of difficulty for language modeling. The same typological feature can appear harmful under one tokenizer or training budget and largely disappear under another. When these factors are removed (e.g., exposure is normalized for sequence-length and vocabulary-allocation effects) the gap between morphologically simpler and richer languages becomes substantially smaller.

\subsection{Lexical Diversity and Vocabulary Size}\label{sec:lexical}

\begin{tcolorbox}[colback=red!2!white,colframe=red!70!black]
Lexical diversity effects reflect tokenization misalignment, not linguistic complexity.
\end{tcolorbox}

Lexical diversity captures how many distinct lexical types (lexemes and multiword
expressions) a corpus contains and how evenly their frequencies are distributed.
In human language acquisition, learning is strongly frequency-driven:
high-frequency forms are acquired earlier, while low-frequency items are acquired
later and remain harder to access~\citep{ambridge2015ubiquity}, reflecting the
long-tailed (Zipfian) distribution of words~\citep{zipf1935psycho}. This
connects lexical diversity to learnability: larger effective vocabularies entail
longer tails of rare items, raising sample complexity for learning word meanings
even if speakers ultimately master them.

Cross-linguistic differences in lexical diversity reflect lexicalization choices
(what is expressed as a single word versus a multiword expression) and
word-formation productivity (derivation and
compounding)~\citep{booij2005compounding, baayen2009productivity}. For instance,
languages differ in how motion events are lexicalized (e.g., encoding manner
versus path in the verb)~\citep{talmy2000cogsem, allen2007manner}. Lexical
diversity is typically measured from word-segmented corpora via type-frequency
distributions, using indices like Type-Token Ratio and its length-normalized
variants~\citep{covington2010mattr, mccarthy2010mtld,
kettunen2014ttr}.\footnote{In corpus linguistics, these indices typically treat
``tokens'' as word tokens in a word-segmented corpus (not subword tokens produced
by NLP tokenizers).}

In multilingual LM analyses, lexical diversity predicts perplexity and transfer
quality~\citep{mielke-etal-2019-kind, pelloni2022sue}. However, perplexity alone
does not reveal which linguistic attributes are
learned~\citep{meister2021beyondperplexity}. Output-side analyses also examine
linguistic diversity in generations: \citet{guo2025benchmarking} evaluate model
outputs along lexical, syntactic, and semantic diversity dimensions and find that
current LLMs fall short of human-level linguistic diversity.

More generally, lexical diversity is a robust predictor of difficulty for LMs:
Head-POS entropy~\citep{dehouck-denis-2018-framework} and raw type counts can
outperform typological features in predicting language modeling
difficulty~\citep{mielke-etal-2019-kind}, and tokenization-sensitive measures
such as Subword Evenness predict cross-lingual transfer and multilingual
perplexity~\citep{pelloni2022sue}. Vocabulary-richness features also predict
GPT-2 perplexity in English and interact with segmentation choices across
typologies~\citep{miaschi2021makes, parra2024morphological}.

However, much of this effect reflects segmentation artifacts: in morphologically
complex languages, frequency-based subwords fragment words into many pieces,
inflating sequence length and reducing effective exposure per unit of semantic
content~\citep{lundin2025tokentax}. When training data is equalized by byte
premium or when tokenization artifacts are otherwise controlled, apparent
lexical-diversity effects weaken
substantially~\citep{arnett2025language}. Lexical diversity, therefore,
challenges LMs mainly under tokenization schemes that misalign with linguistic
structure. Vocabulary size remains a strong predictor, mainly due to segmentation
and data sparsity rather than inherent lexical complexity.

\subsection{Syntactic Features}%: Word Order, Case, and Structural Variation}
\label{sec:syntax}

\begin{tcolorbox}[colback=red!2!white,colframe=red!70!black]
Syntactic features affect modeling difficulty indirectly through interactions with morphology, vocabulary size, and tokenization.
\end{tcolorbox}

Syntactic features describe how languages organize words into phrases and clauses, including word order, case marking, and dependency structure. Syntax and morphology often provide alternative encodings for the same grammatical distinctions: a language may rely more on word order or more on overt marking (case, agreement) to signal roles and relations while preserving overall communicative efficiency~\citep{sinnemaki-2008-tradeoff, lian-etal-2023-communication, levshina2021cross, fedzechkina2017balancing}. In humans, these trade-offs influence which cues must be tracked rather than creating a global difficulty hierarchy.

The evidence on the effects of word order and syntactic variation on language modelling difficulty remains mixed~\citep{mielke-etal-2019-kind, miaschi2021makes}, and syntax is generally less investigated than morphology and tokenization in this context.
%For LMs, syntactic variation affects surprisal and perplexity, but effects are typically smaller and more indirect than morphology or tokenization~. 
Analyses based on typological features typically find that syntactic typology explains less variance in surprisal and perplexity than tokenization or lexical measures, with the largest effects occurring when critical syntactic cues rely on morphemes that subword tokenizers fragment~\citep{mielke-etal-2019-kind}. Case marking illustrates this interaction: under standard BPE, languages with productive case systems show higher surprisal, but morphology-aware segmentation reduces the gap by segmenting case morphemes more consistently ~\citep{park-etal-2021-morphology}, increasing their effective frequency and preserving cues for syntactic roles.
Word order effects are mixed: basic order alone is not a reliable predictor of perplexity~\citep{mielke-etal-2019-kind}, and reducing word-order-specific encoding can improve cross-lingual adaptation~\citep{liu2021wordorder}. Dependency distance metrics or embedding depth could, in principle, affect language modelling difficulty, but to the best of our knowledge, they have not yet been studied in this context.
%Dependency length adds another constraint: longer dependencies increase context requirements and interact with architectures biased toward English-like branching~\citep{gibson-1998-splt, futrell2015large, hewitt2019structural}.

In sum, syntactic differences rarely govern language modelling difficulty when taken in isolation; rather they interact with tokenization artifacts that inflate sequence length or obscure morphological cues~\citep{arnett2025language}. Consequently, syntax-related performance gaps often reflect architectural constraints and English-centric positional heuristics rather than inherent modelling difficulty. 
Cognitively-motivated inductive biases, such as relative position encodings and syntactically informed attention, can mitigate these issues~\citep{shaw2018self, dufter2022position, strubell2018linguistically, kuribayashi2024emergent}, but positional design choices matter and multilingual evidence for newer schemes (ALiBi, RoPE) remains mixed~\citep{ravishankar2021positional, press2022alibi, su2024roformer}.

Overall, syntactic variation shapes cross-linguistic gaps mainly through interactions with morphology, tokenization, and vocabulary size; once normalized, syntax alone explains less of the variance, though it remains important for generalization and cross-lingual transfer.

\subsection{Information-Theoretic Measures}%: Entropy, Compression, and Predictability}
\label{sec:information}

\begin{tcolorbox}[colback=red!2!white,colframe=red!70!black]
Entropy differences largely capture representational choices, like word length or morphological encoding, rather than the intrinsic learnability of a language.
\end{tcolorbox}

Information-theoretic metrics quantify predictability and redundancy, but they also reflect morphology, orthography, and other representational choices rather than pure learnability. Some potentially informative metrics remain difficult to define or measure, leaving room for future work. Information-theoretic measures quantify predictability and redundancy: \emph{entropy} captures average uncertainty, \emph{surprisal} measures the negative log probability of an observed unit, and \emph{compression rate} approximates achievable code length under efficient encoding. These metrics provide a principled way to compare languages in terms of predictability and coding efficiency, linking cross-entropy in LMs to fundamental data statistics~\citep{shannon1948mathematical}. In human processing, surprisal theory formalizes the connection between predictability and cognitive difficulty~\citep{hale-2001-surprisal, smith-levy-2013-logarithmic}.

A central insight from psycholinguistics and quantitative linguistics is that languages maintain stable information rates through compensatory trade-offs. Spoken languages converge on near-constant bits-per-second rates~\citep{coupe2019different, jaeger2010redundancy}, and morphologically rich languages exhibit higher per-word entropy because they encode more information per word~\citep{bentz2016entropy, koplenig2025tradeoff}. 

Large-scale studies show systematic differences in entropy at the character and word levels, balanced by structural features like word length~\citep{koplenig2025tradeoff}. This reflects \textsl{Uniform Information Density} (UID), where languages spread information to keep local surprisal relatively stable~\citep{jaeger2010redundancy, levy-jaeger-2007-uid}, though UID might not be a universal law~\citep{meister-etal-2021-revisiting-uid}.

For LMs, entropy interacts with tokenization and sampling: high-entropy sequences require more data, and token-based budgets can exaggerate difficulty when scripts or tokenizers inflate sequence length. Byte-inefficient scripts and fragmented tokenization can inflate apparent entropy without adding semantic content~\citep{rust2021good}.
At the human-processing level, LM surprisal estimates can predict reading times across multiple languages, suggesting that surprisal is a useful--but imperfect--proxy for cognitive difficulty in cross-linguistic comparisons~\citep{levy2008expectation, goodkind-bicknell-2018-predictive, hollenstein-etal-2021-multilingual-lms-reading, devarda-marelli-2022-surprisal-across-languages, wilcox-etal-2023-surprisal-11-languages, siegelman-etal-2025-meco-wave2}.

Controlling for encoding efficiency and tokenization substantially reduces cross-linguistic surprisal gaps and narrows perplexity differences, indicating that part of the observed entropy variation reflects representation and sampling confounds~\citep{arnett2024bit, rust2021good, foroutan2025paritybpe, tsvetkov2024parity}. However, perplexity remains an imperfect proxy for downstream performance: low perplexity can coexist with weak robustness, particularly in low-resource settings~\citep{luitel2025perplexity, gurgurov2025small, zhuang2025cute, liu_2023_same_loss, lourie_2025_scaling_unreliable}.

Compression-based metrics provide architecture-independent baselines by evaluating predictability at fixed representational units. Bits-per-character/byte (BPC) estimates cross-entropy per character/byte, reducing reliance on subword tokenization and enabling comparisons that align with LM perplexity and transfer~\citep{de2024measuring, tsvetkov2024parity}. However, BPC is encoding-sensitive: UTF-8 byte premiums and script granularity distort comparisons even after byte normalization~\citep{arnett2024bit, moon2025bitlevel, foroutan2025paritybpe, deletang2024language}.% Recent work formalizes the language modeling-compression link, highlighting BPC as a complementary lens~\citep{deletang2024language}.

In sum, information-theoretic differences reflect language encoding choices rather than inherent learnability, and normalizing for density, byte length, or morphemes reduces many cross-linguistic gaps.

\subsection{Typological Distance}\label{sec:typology}

\begin{tcolorbox}[colback=red!2!white,colframe=red!70!black]
Typological diversity can cause difficulty in shared-parameter \& -vocabulary settings.
\end{tcolorbox}

Let's now move from single-language difficulty to cross-linguistic transfer, where patterns learned in one language can improve performance in others. 

First, languages differ in many ways; this  diversity represents alternative solutions to similar communicative constraints~\citep{bickel2013distributional, comrie1989language, j19-3005}. We can measure the difference between a pair of languages  via their
typological distance, which captures similarity in grammar (syntax, morphology), lexicon (cognates, word choice), and phonology, or via genealogical relatedness, which reflects shared ancestry.

In human L2 acquisition, linguistic distance predicts attainment and learning difficulty~\citep{chiswick-miller-2005-linguistic-distance,isphording-otten-2014-linguistic-barriers, schepens-vanhout-jaeger-2020-constraints}. 
Similar results have been suggested in models. Early multilingual models show that shared vocabularies bias representations toward related languages~\citep{pires2019multilingual,wu2019emerging}, with mBERT organizing languages along genealogical lines~\citep{rama_beinborn_eger_2020_probing_mbert}.

At a finer level, WALS-based similarity~\citep{haspelmath2013wals} predicts transfer quality beyond raw resource size~\citep{lin2019choosing}, with features like word order and head direction particularly predictive~\citep{k2020cross, blaschke_fedzechkina_terhoeve_2025_similarity_transfer}. Tokenization-based diagnostics like Subword Evenness~\citep{pelloni2022sue} and information-theoretic metrics like Information Parity~\citep{tsvetkov2024parity} can predict cross-lingual transfer.
Vocabulary overlap can sometimes predict positive transfer but sometimes be detrimental, depending on the exact task~\citep{limisiewicz_etal_2023_vocab_overlap}. For example
\citet{kallini_etal_2025_false_friends_overlap} found that even  vocabulary overlap of semantically unrelated words can be useful.

 At larger scales, the \emph{curse of multilinguality} refers to declining per-language performance as more languages share parameters~\citep{conneau-etal-2020-unsupervised}, with low-resource and typologically distant languages suffering most~\citep{lauscher2020zero}. Typological distance amplifies interference and exacerbates vocabulary fragmentation. Controlled studies show that adding related languages improves low-resource performance, but may surprisingly hurt performance on high-resource languages~\citep{chang2024multilinguality}. Gradient conflicts are common when distant languages are trained jointly~\citep{wang2020negative}.

In summary, shared-parameter training with similar languages can help, but can also  induce interference as typological diversity grows. Modular approaches that allocate language-specific capacity reduce conflict while preserving positive transfer~\citep{pfeiffer2022lifting, blevins2024breaking}.

\subsection{Summarizing Linguistic Properties}

Across features, many performance gaps arise from mismatches between linguistic structure and modeling choices rather than intrinsic language difficulty. Tokenization and encoding can fragment cues and lengthen sequences, sampling can create unequal exposure, and shared-parameter training can cause negative transfer when typological diversity exceeds capacity. These factors also confound evaluation: low perplexity does not guarantee robust downstream performance, especially in low-resource settings. When segmentation, encoding, and exposure are normalized, many apparent cross-linguistic gaps shrink, showing that current modeling paradigms, not linguistic diversity itself, drive much of the disparity.

\section{Design Implications}

% \cnote{Maybe split to two sections? "Designing multi lingual LLMs" and "Reporting checklist for multilingual LLMs"}

These findings motivate design implications for tokenization, sampling, architecture, evaluation, and corpus construction; we focus on interventions most directly supported by the surveyed evidence.

\subsection{Tokenization: From Frequency-Based Segments to Linguistically Informed Units}\label{sec:tokenization}

Tokenization is one of the main design levers through which cross-linguistic disparities are either amplified or mitigated in multilingual language models. Frequency-based subword algorithms such as BPE and WordPiece often fragment morphemes and disproportionately disadvantage multi-byte scripts, inflating sequence length and compute cost while obscuring linguistically meaningful units~\citep{park-etal-2021-morphology,ali-etal-2024-tokenizer,land2025script,petrov2023language,arnett2024bit,lundin2025tokentax}. Across studies, segmentation quality explains a substantial share of cross-linguistic performance differences, and morphology-, script-, and encoding-aware methods improve performance and efficiency across languages~\citep{limisiewicz2024myte,asgari2025morphbpe,mager2022bpe}.

Assessing tokenization quality remains nontrivial. Common diagnostics include compression, sequence length, corpus token count, vocabulary-balance measures, and related distributional metrics, but improvements on these measures do not always translate directly into better downstream performance~\citep{schmidt2024tokenization,zouhar-etal-2023-tokenization,goldman2024unpacking,dagan2024getting,galle-2019-investigating}. Tokenization should therefore be evaluated jointly in terms of efficiency, downstream utility, and parity across languages.

The surveyed work points to three broad intervention families. First, morphology-aware and language-adaptive tokenization can better preserve recurring morphemes and improve parity across languages~\citep{asgari2025morphbpe,foroutan2025paritybpe,ahia2024magnet}. Second, alternative byte encodings can reduce disparities that arise when standard UTF-8 and shared-vocabulary BPE allocate capacity unevenly across scripts or create partial-byte artifacts~\citep{limisiewicz2024myte,land2025script}. Third, tokenizer-free and character-level models can reduce script-specific tokenization penalties by avoiding fixed subword vocabularies altogether and reduce cross-lingual performance gaps, though typically at the cost of longer sequences and higher compute~\citep{pagnoni-etal-2025-byte,clark2022canine}.

\noindent \textcolor{red!70!black}{\textbf{Implication 1:}} Treat tokenization as a first-class multilingual design choice rather than a fixed preprocessing step. Prefer morphology-aware, script-aware, or language-adaptive tokenizers that better preserve meaningful units and allocate vocabulary capacity more evenly across languages. Consider alternative byte encodings to reduce systematic disadvantages introdcued by standard UTF-8 encoding. Tokenizer-free models can further reduce cross-lingual disparities, at the cost of longer sequences and higher compute.

\subsection{Data Sampling and Byte Normalization}\label{sec:sampling}

Token-based sampling penalizes byte-heavy scripts, whereas byte-normalized sampling narrows gaps~\citep{arnett2024bit, wei2021bytelevel}, motivating sampling strategies that target semantic exposure rather than raw token counts (e.g., UniMax, byte-premium scaling; ~\citealp{chung2023unimax, chang2024goldfish, he-etal-2025-multilingual-scaling-laws}).

\noindent \textcolor{red!70!black}{\textbf{Implication 2:}} Pretraining should use \textsl{byte-normalized}, \textsl{information-normalized}, or \textsl{morpheme-normalized} sampling for equal semantic coverage across languages. 
Data balancing should reflect linguistic diversity rather than corpus availability, correcting for segmentation bias, type proliferation, and script inefficiency.

\subsection{Beyond One-Size-Fits-All Benchmarks}\label{sec:evaluation}

Current multilingual benchmarks often conflate linguistic difficulty with tokenization artifacts or dataset size. Perplexity is sensitive to tokenizer choice, whereas character-, morpheme-, and byte-level metrics provide more robust comparisons~\citep{tsvetkov2024parity, kanjirangat2025biases}. Tokenizer-quality diagnostics and standardized reporting help disentangle measurement bias from true modeling capability~\citep{chelombitko_etal_2024_qtok, bender_friedman_2018_datastatements}.

Cross-linguistic syntactic challenge suites like CLAMS offer controlled tests of generalization and reveal consistent gaps between monolingual and multilingual models~\citep{mueller-etal-2020-cross}. For morphology, community benchmarks (e.g. SIGMORPHON) provide fine-grained metrics that complement perplexity-based ones~\citep{cotterell2017conll}.

Probe-based evaluations show representational disparities, such as weaker subject/object identification in case-rich languages when models are trained on fixed word order~\citep{papadimitriou_etal_2021_deep_subjecthood}, motivating typology-aware competency assessments.

\noindent \textcolor{red!70!black}{\textbf{Implication 3:}} Evaluation should use linguistically informed metrics and typology-aware probes beyond subword perplexity. Benchmarks should disaggregate performance by morphology, script, and word order to avoid masking inequities.

\subsection{Balanced Corpora and Pretraining}
\label{sec:balanced_pretraining}

Pretraining corpus choice strongly shapes multilingual performance. Web data overindexes English and underrepresents high-vitality languages, correlating poorly with global populations~\citep{dunn2020corpus, dunn2020mapping, mehmood2017commoncrawl, mor2025globalvillage, joshi-etal-2020-diversity, khanna2025invisible, bella2023bridging}. Corpus composition often tracks speaker counts over linguistic diversity, while data statements support accountable multilingual reporting~\citep{bender_friedman_2018_datastatements}.

Multilingual corpora favor Indo-European languages and underrepresent complex morphology, minority scripts, and small populations. Balancing must account not only for token counts but also for linguistic density: information per token, morphological productivity, and rare-form distributions.

Resources such as UniMorph and high-coverage dependency treebanks can support typology-aware evaluation of coverage, even without direct training supervision~\citep{nivre2020universal}. These findings motivate moving beyond a single monolithic model: leveraging language similarity or tailoring components to typologically related clusters can boost learning for low-resource languages without forcing uniform representations~\citep{malkin-etal-2022-balanced}.

\noindent \textcolor{red!70!black}{\textbf{Implication 4:}} Corpus design should explicitly encode linguistic diversity by accounting for representational efficiency and linguistic density, ensuring that languages with high morphological or typological variation receive equivalent \emph{semantic coverage}, not merely equivalent token counts.

% \cnote{Do we actually need one model for all languages? maybe dynamically select similar languages for boosting learning for a low resource language? https://arxiv.org/pdf/2205.04086}

% \subsection{Linguistic ``Difficulty'' as a Constraint}
% \label{sec:design_constraint}

% Ultimately, ``difficulty'' in multilingual LMs reflects mismatches between model assumptions and linguistic structure. Cross-linguistic disparities shrink when units are preserved, sampling normalized, and architectures interference-free~\citep{jaeger2010redundancy, gibson2019efficiency, bickel2013distributional}.

% \noindent \textcolor{red!70!black}{\textbf{Implication 5:}} Treat linguistic diversity as a design constraint, ensuring model components respect language structure. This framing supports typology-aware models that respect linguistic diversity rather than reinforcing English-centric biases.

\section{Conclusions and Future Work}

Multilingual performance is shaped less by inherent linguistic complexity than by design choices: tokenization, data allocation, and interference-aware training.
Future work should explore language-adaptive strategies: predicting data and capacity needs per language, designing curricula that prioritize transfer from related languages, and developing architectures that dynamically allocate resources across typologically distinct languages. Truly low-resource and endangered languages require innovative approaches under scarcity.

Evaluation must also evolve: metrics should reflect cross-linguistic differences in task difficulty while capturing fairness and accessibility. Aligning model inductive biases with human learning can guide more robust multilingual NLP.
 
By embracing \textbf{linguistic diversity as a design principle}, we can build models that are more adaptable, equitable, and capable of supporting the full spectrum of the world’s languages.

\section{Limitations}
Despite our analysis of multilingual performance, several limitations warrant consideration. First, our work focuses primarily on representation- and architecture-driven factors (tokenization, encoding, shared parameters) and does not fully capture other potential sources of difficulty, such as pragmatic, discourse-level, or sociolinguistic phenomena, which may affect real-world usage.

Second, most of our empirical insights rely on pretrained models and standard evaluation datasets, which may underrepresent truly low-resource or endangered languages. Data sparsity, orthographic variation, and non-standardized corpora in such languages could yield patterns not observed in higher-resource languages.

Third, while we consider cross-linguistic typology, our analysis is largely English-centric in architecture and benchmark design, which may bias conclusions about syntax, word order, and positional encoding effects.

Fourth, information-theoretic measures capture correlations with morphology and orthography rather than intrinsic learnability. Metrics for hierarchical structure, discourse-level predictability, or multimodal signals remain underexplored, leaving important aspects of language modeling outside our current framework.

Finally, despite our thorough literature survey, it is possible that relevant works were overlooked. We welcome pointers to such papers to keep this survey up to date. 

Addressing these limitations in future work will be crucial for building truly language-adaptive, equitable, and robust multilingual models.

\textbf{AI usage:} The paper used AI assistance for rephrasing, for finding additional relevant papers, and occasionally for summarizing them.

% \newpage

\bibliography{custom}

\appendix

% \section{Example Appendix}
% \label{sec:appendix}

\end{document}